# Title of the article:

Quality of Data in Machine Learning

# Authors:

Antti Kariluoto, Arto Pärnänen, Joni Kultanen, Jukka Soininen, and Pekka Abrahamsson

# NOTES:

- This is the author's version of the work.
- The definite version was presented in International Workshop on Data Quality for Intelligent Systems (DQIS), which was a co-located event of QRS 2021 (The 21st IEEE International Conference on Software Quality, Reliability, and Security).

# Quality of Data in Machine Learning

Antti Kariluoto
University of Jyväskylä
antti.j.e.kariluoto@jyu.fi
Arto Pärnänen
TCD Consulting and Research Oy
arto.parnanen@tcdcon.com
Joni Kultanen
University of Jyväskylä
joni.m.kultanen@jyu.fi
Jukka Soininen
TCD Consulting and Research Oy
jukka.soininen@tcdcon.com
Pekka Abrahamsson
University of Jyväskylä
pekka.abrahamsson@jyu.fi

*Abstract*— **A common assumption exists according to which machine learning models improve their performance when they have more data to learn from. In this study, the authors wished to clarify the dilemma by performing an empirical experiment utilizing novel vocational student data. The experiment compared different machine learning algorithms while varying the number of data and feature combinations available for training and testing the models. The experiment revealed that the increase of data records or their sample frequency does not immediately lead to significant increases in the model accuracies or performance, however the variance of accuracies does diminish in the case of ensemble models. Similar phenomenon was witnessed while increasing the number of input features for the models. The study refutes the starting assumption and continues to state that in this case the significance in data lies in the quality of the data instead of the quantity of the data.**

***Keywords- machine learning; data quality; classification;***

## I. INTRODUCTION *(HEADING 1)*

The question of pouring data to machine learning models in hopes of gaining better results is a somewhat prevailing idea in the academia and industries. However, sometimes it helps to add more data while in some other cases the increment of data does not help. This study tackles this problem by experimenting with student performance data from Finnish vocational educational institution and comparing different machine learning algorithms under the task of classifying the students. The used machine learning models were decision tree (DT) and random forest (RF) that are typical classifying algorithms found, for example in the explainable artificial intelligence (XAI) literature and other fields of the artificial intelligence research. The metrics used in the study were accuracy, roc auc score, precision, recall, and performance.

The authors were interested of the relationship between data and machine learning model's learning. Therefore, the following research questions were formulated. First, what happens to the model's performance when more data is used in the learning phase of the model? Second, will increasing the number of input features improve the model's performance? Our research's contribution is giving primary empirical contributions based on the experiment done on the collected unique data.

## II. Dataset

This article attempted to answer the raised research questions by implementing two experiments with anonymized Finnish vocational student data. This section explains the data.

### A. *Description of the data*

The batch of data used in this experiment consisted of the features listed in the column Feature in table 1. The features Study day, HOKS events (i.e., events of personal competence development plan), Absences, On-the-job learning, Days of on-the-job learning, Course completions, Failed course completions, Competency tests, Disciplinary actions, and Days of temporary suspension had daily cumulative values which were calculated for each student every day even though not all of the events could happen to each student every day and depending of the unit of the feature not every feature have an event every time. E.g., days from the beginning of the student's studies in the vocational school measured in days updated differently compared to student's age measured in years or to the number of absences that were counted as events. Feature Home town indicated whether or not the student was originally from the same town as in where the educational institution was situated. Feature Target degree had the name of the student's chosen study degree, while the feature Degree type indicated whether or not the student had chosen a full degree course or some partial performance. Education objective defined if the course was part of the requirements meant for the students to pass before obtaining a vocational competence. Feature Student had uuids for anonymized students. The feature start date was a set date value when the student began their studies and End date had date values for the graduated or dropped out students while no date value was included for students who were still present in their studies or had their studies temporarily suspended. The feature Final study state indicated the students final situation in the educational institution, i.e. had they graduated or dropped out from their studies.

Other columns of the table 1 are group_type, data_type, fillna_value, and use_trained_encoder. The group_type column tells the system which of the features should be utilized as input or output to the ML models and which of the data is metainformation. The metainformation is not used with the machine learning models themselves but with the processing of the data and code flow. The data_type column indicates what the data type for the feature's values ought to be. When the value is an integer number system is to use the 32-bit representation of the values while the value "object" in the same column tells the system the values of the feature are of type string. The fillna_value column tells the system which value to use as replacement for missing values, i.e., to replace any missing or null values in numerical feature columns with -1 and to use string "<Unknown>" for any features that are of string (or in Python "object") type.

TABLE I. DESCRIPTION OF INPUT AND OUTPUT FEATURES.

| Feature | Feature (english) | group_type | data_type |
|---|---|---|---|
| Paiva | Day | input-multiple | int32 |
| RahoituksenKoodi | Funding code | input-multiple | int32 |
| HOKSTapahtumat | HOKS events | input-multiple | int32 |
| Poissaolot | Absences | input-multiple | int32 |
| Tyossaoppimisia | Learn on the job | input-multiple | int32 |
| Tyossaopp_paivat | Days of learning on the job | input-multiple | int32 |
| Suorituksia | Accomplishments | input-multiple | int32 |
| Suorituksia_hyl | Failed accomplishments | input-multiple | int32 |
| Nayttoja | Exhibits | input-multiple | int32 |
| Kurinpitoja | Disciplinary actions | input-multiple | int32 |
| ValiaikaisenKeskeytyksenPaivat | Days of temporary suspension | input-multiple | int32 |
| Ika | Age | input-singular | int32 |
| Kotikunta | Home town | input-singular | object |
| Tutkinto | Degree | input-singular | object |
| Tutkintotyyppi | Degree type | input-singular | object |
| KoulutuksenTavoite | Education objective | input-singular | object |
| Opiskelija | Student | metainformation | object |
| AloitusPaivamaara | Start day | metainformation | object |
| LopetusPaiva | End day | metainformation | object |
| Tila | Final study state | output | object |

## III. Experiment design

*1) Machine learning models*

The following machine learning algorithms were used or considered in the experiments due to their prior suitability for classification tasks.

*a) Decision tree*

According to [4] comprehensive decision tree classifier survey, the decision tree is used to split the problem to several simpler questions that might be easier to solve. This is experienced as decision tree (DT) algorithm forming a set of directed nodes which begin from the root node and continue to branch into two different nodes at each level of the DT's decision path. Each node except leaf nodes (the last nodes) has a Boolean condition the data must clear in order for the algorithm to progress to the next node in the decision path. The conditions are related to the input features and the driving force of the continuation of the algorithm itself is determined typically either "Gini" or "entropy" information gain metrics.

*b) Random Forest*

Random forest (RF) is an ensemble algorithm which creates, trains, and collects the results of many decision trees internally and outputs a single value as an outcome of the task. It can be said to improve its base algorithm because the superposition of the base models' results can help to balance out some of the biased answers. However, the bias can exist in the final model because it creates inner models on random beginning values similarly to decision trees and the data might have similar suborder within other inner models as well. According to [2], the algorithm is resilient when it comes to imbalanced data sets and noise in the data besides being relatively fast to train and use to do predictions.

*c) Support Vector Machine*

Support vector machine is a classical machine learning algorithm which is often used in cases where the multidimensional data needs to be classified between two different choices with some ambiguous decision boundary. The classification errors of the decision boundary are minimized utilizing the training data which allows the model remain more general than for example a decision tree. [1]

Although the SVM classifier model was a candidate model in the experiments the trial runs of the experiments code revealed that it could not be used for the experiments as they were designed. This was due to the fact that after training data records number grows to tens of thousands the training time of the model quadruples. [3]

*2) Techniques and metrics*

The following techniques were used in the pre-processing of the data before training the machine learning models. Balancing by down sampling, primary component analysis (PCA), and scaling of input features for the PCA.

The following metrics were used to measure the performance of the trained ML models: accuracy, Receiver Operator curve Area Under the Curve (ROC AUC),

Jaccard, recall, and precision.

### B. *Description of the set-up*

Two code scripts were written in Python 3.8 for each experiment, respectively. The experiment scripts were run in Ubuntu 20.10 guest virtual machine with Virtualbox 6.1 system with at most 10000mbit of allocated RAM memory and 80 Gt of storage for the guest operating system. The host computer was a spare Windows 10 PC built for virtual reality use for the StartUpLab of Faculty of Information Technology of University of Jyväskylä.

### C. *Performing the experiment and results*

Before doing the experiments, it was necessary to pre-process the batch of data. First data was cleaned by removing students whose final study state was not either graduated or dropped out. Second, utilizing the end day values the students were grouped by their final year, and the finished students' records from 2018 and 2019 became the training and testing data while the 2020 and 2021 records were set aside as validation data. Both data sets were then written to their separate csv files. The table 2 shows the number of finished students per each year.

TABLE II. NUMBERS OF STUDENTS WHO GRADUATED OR DROPPED OUT FROM THEIR STUDIES BETWEEN 2018 AND 2021.

| Year | # of Students | Testing and training, or validation |
|---|---|---|
| 2018 | 1099 | Testing and training |
| 2019 | 942 | Testing and training |
| 2020 | 1053 | Validation |
| 2021 | 95 | Validation |

For all the experiments it was necessary to also generate the distinct and unique values for each nonnumeric input and output features. Thus, this generating code was included in the beginning of each experiment script and the label encoders were fitted with these values before the actual experiments code statements. Each label encoder was fitted for one feature, respectively. Algorithm was later tweaked to include the default value (i.e., "<Unknown>") for missing or unforeseen values.

*1) Experiment 1 introduced – addition of data*

The experiment one attempts to answer the research question 1. "what happens to the model's performance when more data is used in the learning phase of the model?"

The script of the experiment 1 continues from the building of the fitted encoders to load the training and testing data set and divide it randomly into two data frames (i.e., training and testing data frames) by the students' unique student ids. The testing data frame has 20% of the unique students ids of the original data frame. Then the validation data is loaded from its csv file. In the main loop the percentage based sampling is done to both the training data set and the testing data set. In other words, it is selected how large of a percentage of all the training and testing data's records are taken. However, only the training data is balanced by the students' final study state. The training, testing, and validating the machine learning models are all done within the second loop. The Scikit-learn Python library was used to implement the ML models. The results of the models' testing and validation are collected and saved to a csv file at the end of the script. The pseudo code of the experiment one is shown in the Fig. 1.

The following settings were used: data size ranged from 10 percent of all the records to total 100 per cent of the records with a step size of 2 percent units. The number of times the innermost loop iterates was set to 5 effectively training, testing, and validating new ML models five times with the selected data size.

*2) Experiment 2 introduced - addition of dimensions to data*

The experiment two attempts to answer the research question 2: "will increasing the number of input features improve the model's performance?"

The experiment 2 script begins by creating the fitted encoders and then performs the primary component analysis (PCA) on the full list of input features using the training and testing data. The results of the PCA are used to determine the order of significant features of the data. This ordered list of features is then used for selecting the input columns to use when training the machine learning models in a loop. The rest of the code trains, tests, and validates models similarly as it is done in the code loops of experiment 1. At each iteration of the outermost loop the ordered list of features is reduced by the removal of the last element of the list (i.e., the least significant feature or column name is dropped from the list). The Fig. 2 shows the pseudo code of the experiment two.

The following settings were used within the script: the list of features had the maximum number of unique column names obtained with utilizing PCA. Data size ranged from 10 per cent of all the records to total 16 per cent of the records with a step size of 2 percent units. K or the number of times the innermost loop iterates was set to 4.

## IV. The Results

### A. *Experiment 1 – addition of data*

Executing the experiment 1 script led to the training, testing, and validating of 460 models divided evenly between decision tree models and random forest models. The figures 3-7 display the results of both model types against the testing data in the same images thus allowing to compare their metrics and behaviour. In the figure 1, both the models' accuracies are well above 80 percent. Accuracies also seem to be increasing when more records of data have been added to the training and testing of the models. The random forest classifiers gain systematically higher accuracy values than the decision tree classifiers, and these accuracies variance tends to diminish when the data size grows.

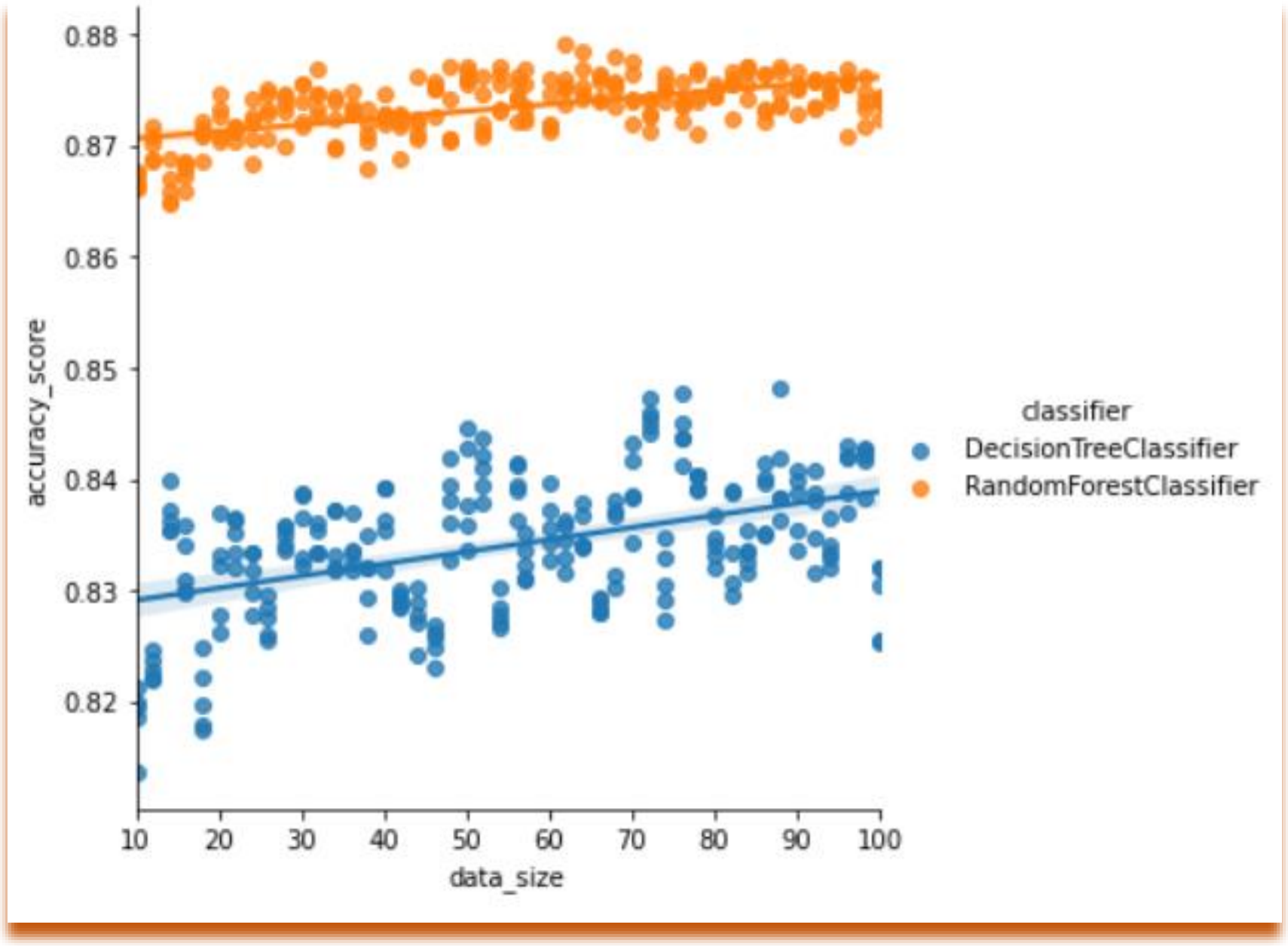

Figure 1. . Accuracy of the ML models of the experiment 1.

The ROC-AUC score is lower than the accuracy score. In the fig. 2, the models give inconclusive results whether or not the increase of data in the training and testing phase actually increases the models' performance.

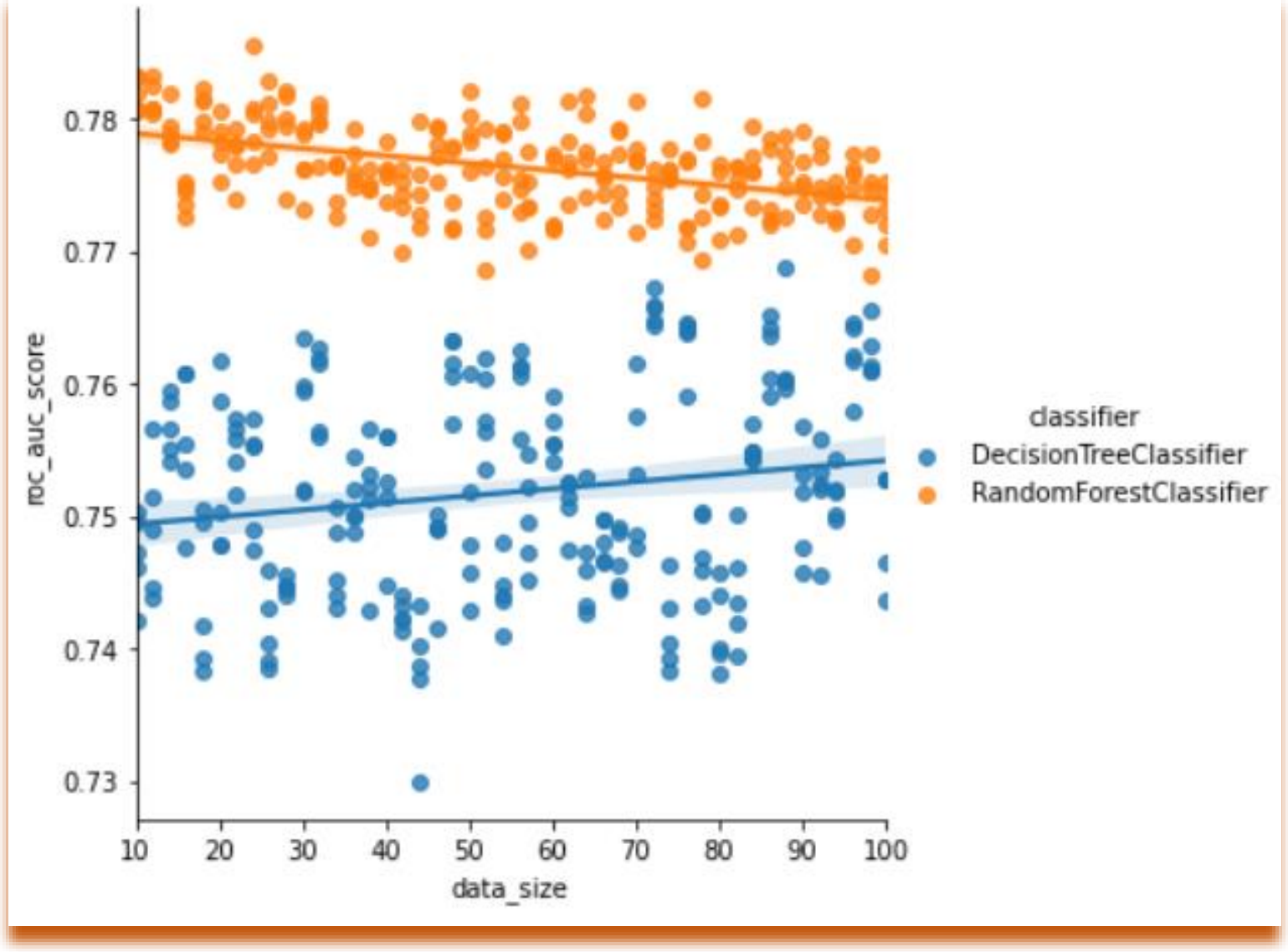

Figure 2. ROC-AUC score of the ML models of the experiment 1.

In fig. 5 are the Jaccard scores for these models. The scores are low and authors' prior understanding is that the Jaccard score metric tend to give lower scores than the other selected metrics in this study. In this case the decision tree models' scores are lower than the random forest models' scores and they have also greater variance despite the increase of data records.

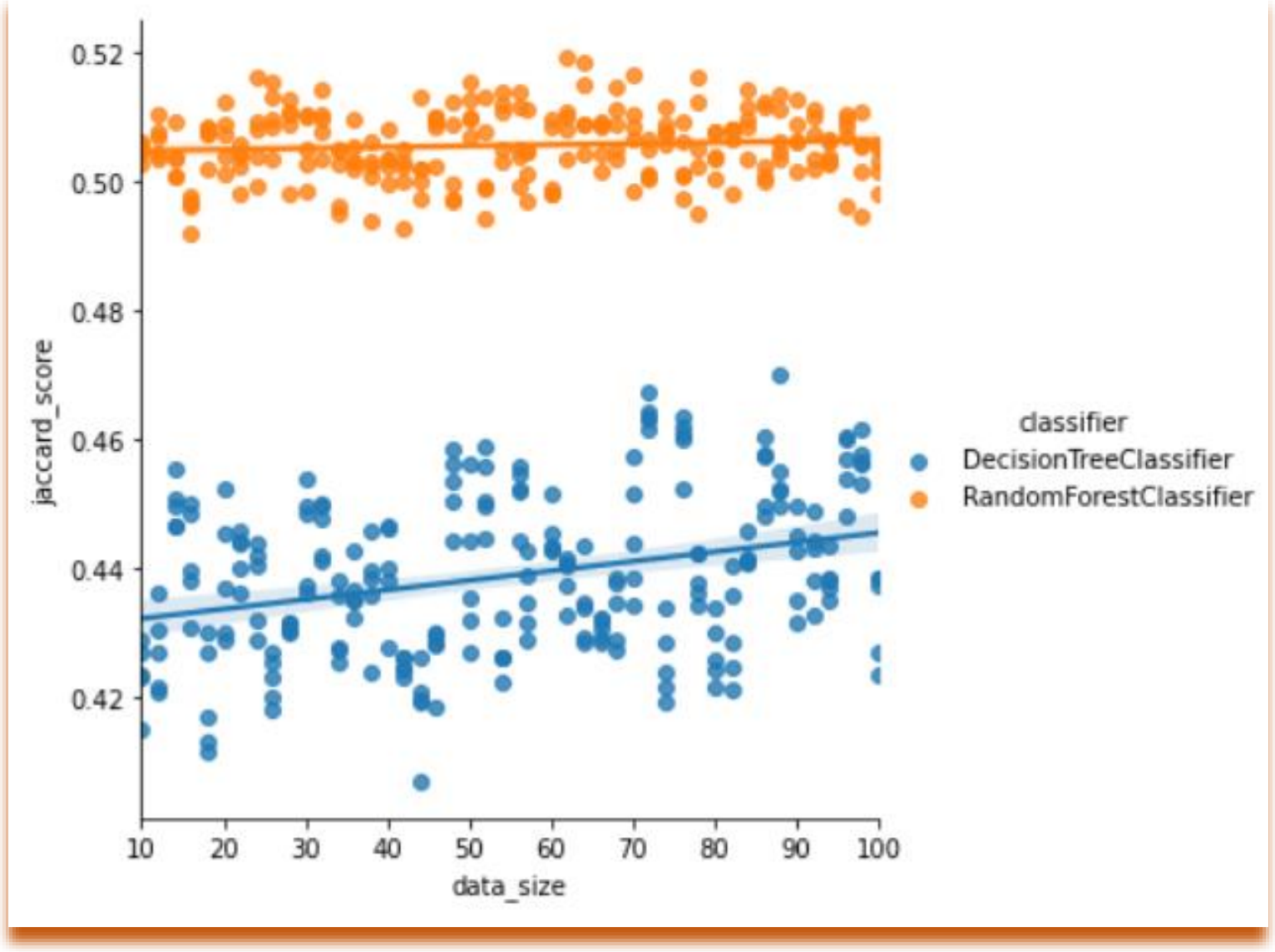

Figure 3. Jaccard score of the ML models of the experiment 1.

In the fig. 6, the models are both giving lessening recall score values when the number of training and testing records grows. Decision tree models' results vary from 57 percent to nearly 64 percent with no sign of diminishing variance. While the same is true with the random forest models their variance is definitely smaller.

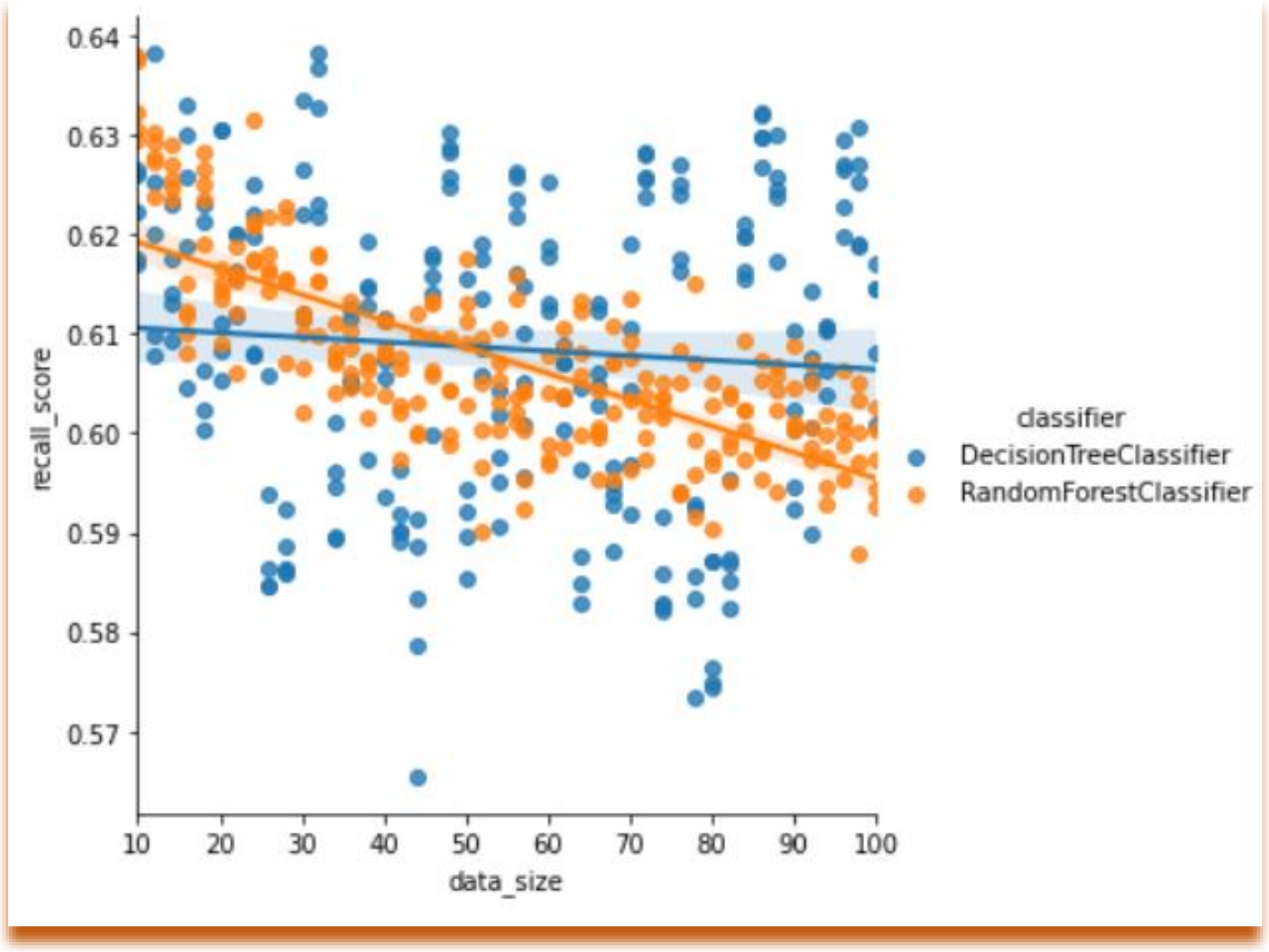

Figure 4. Recall score of the ML models of the experiment 1.

In the Fig. 7 the precision score of the models are shown. Both model type's instances present that the test results improve with the increase of data. The variance of each is smaller than in the case of the recall score.

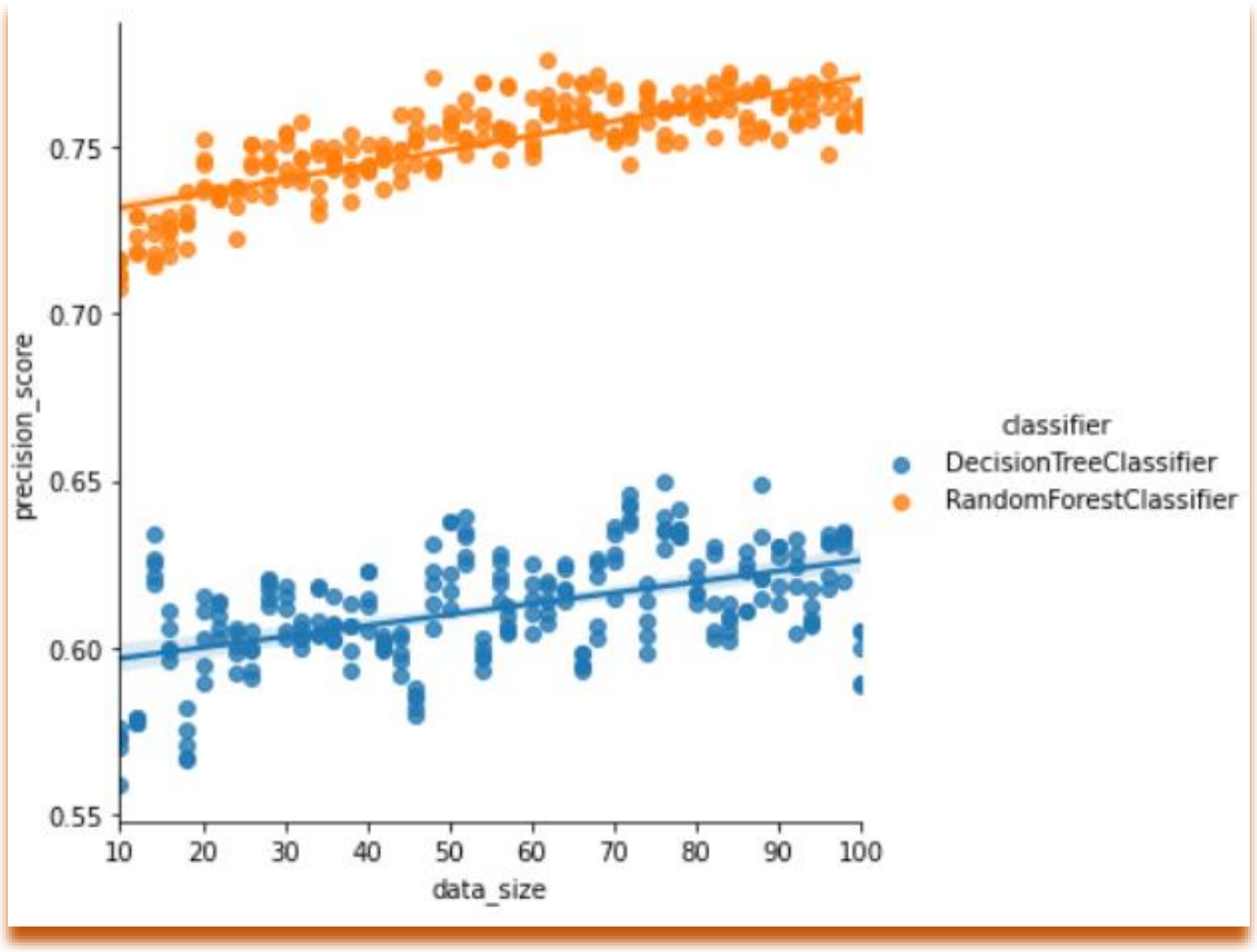

Figure 5. Precision score of the ML models of the experiment 1.

The same models' results with the validation data are shown in the figures 8-11. In the fig. 8 both the random forest models and decision tree models are improving when the number of records has grown in the training data. The random forest models have the highest accuracy value of 97 percent when all of the data available for training the models has been used. The random forests' variances have not grown significantly. While the decision tree model did achieve high accuracy results the variance of those scores fluctuated when approximately half of the data records were used in training.

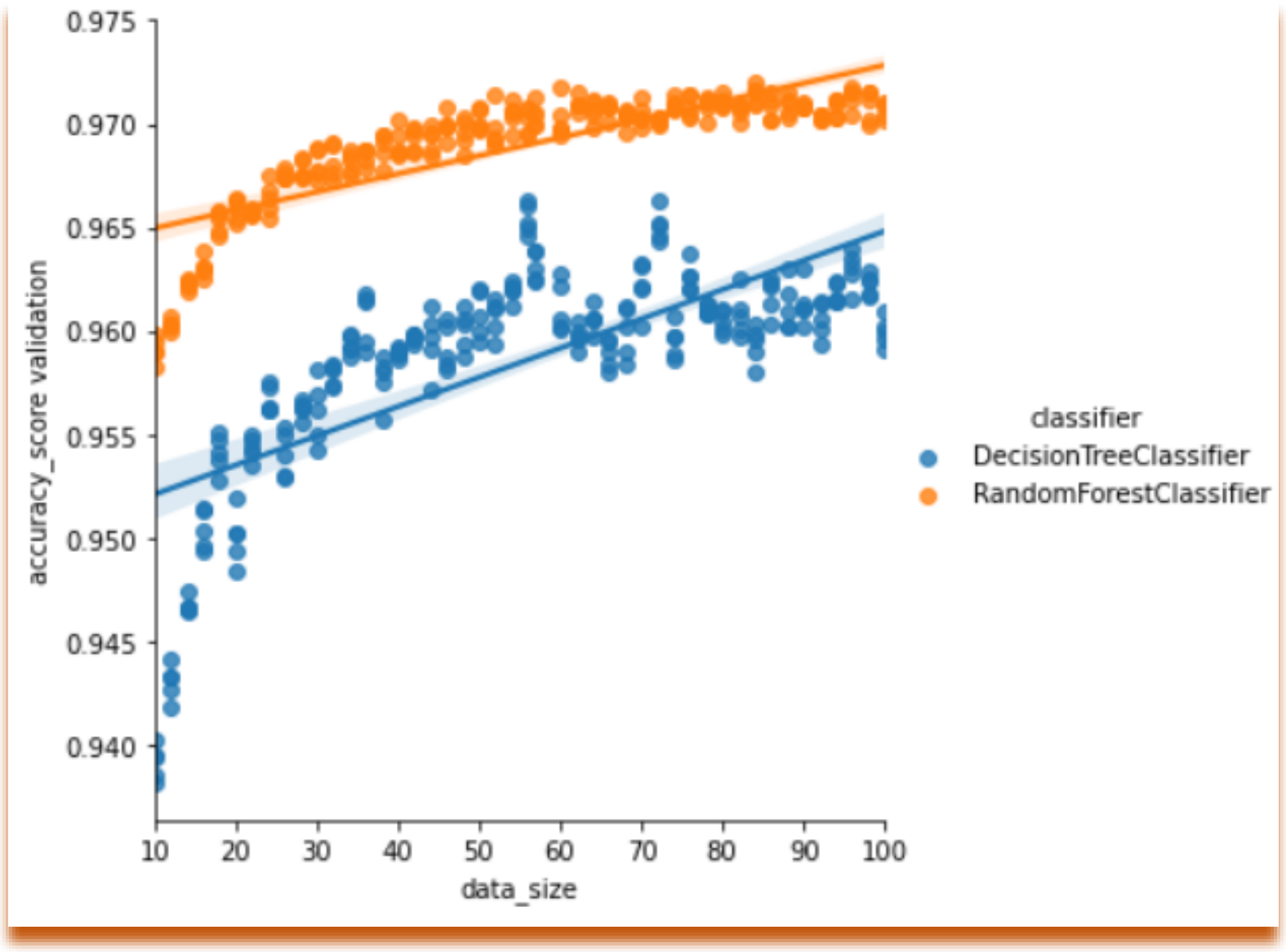

Figure 6. Accuracy score of the ML models of the experiment 1 (validation).

In the Fig. 9 the models are presented with their scores of the ROC-AUC metric. Both models have performed well, and their trend lines are positive. However, the values are lower compared with the accuracy metrics' scores.

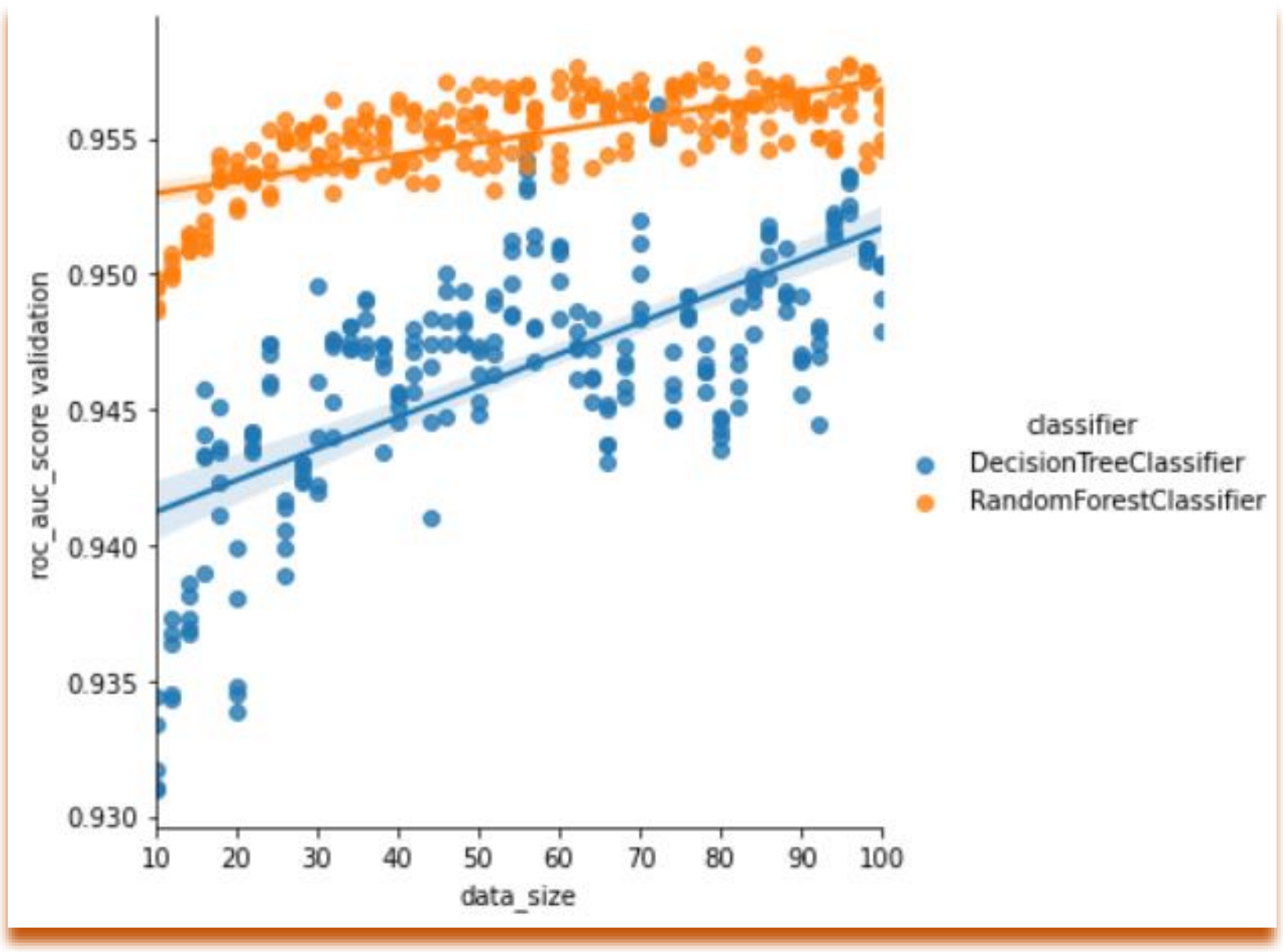

Figure 7. ROC-AUC score of the ML models of the experiment 1 (validation).

In the Fig. 10 the Jaccard score of the models is both high and the models' trendlines are increasing. The results are inconclusive when comparing with the Jaccard score of the testing data.

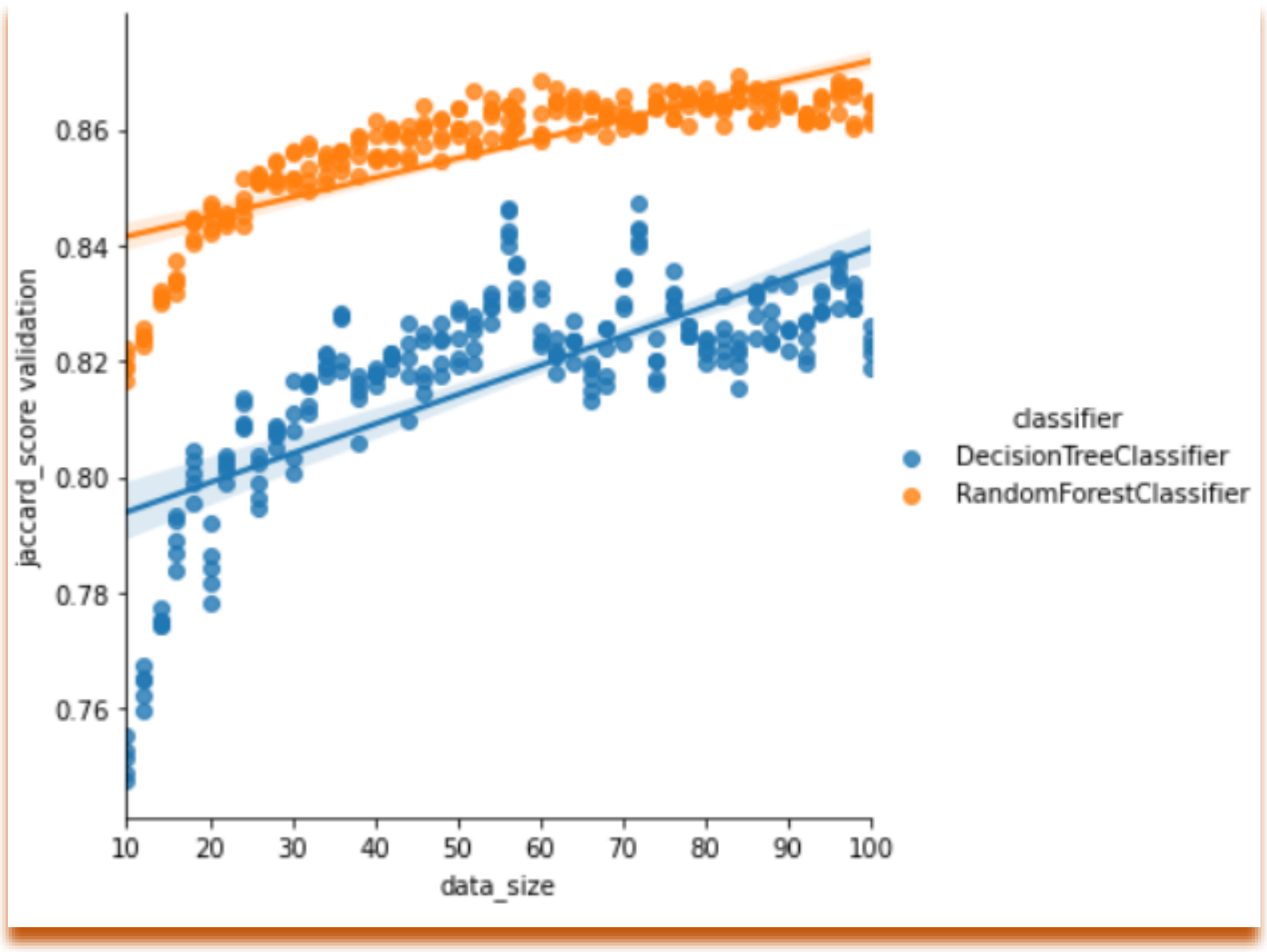

Figure 8. Jaccard score of the ML models of the experiment 1 (validation).

The Fig. 11 represents the recall scores of the validation phase of the models. Results of the metric are high for each model with the lowest being a little less than 91.0 percent (decision tree model at 20 percent of the records in use). However, the models' results are yet again inconclusive since the decision tree classifiers' scores are increasing while the random forest models' results are decreasing when more of the records have been utilized to train the models.

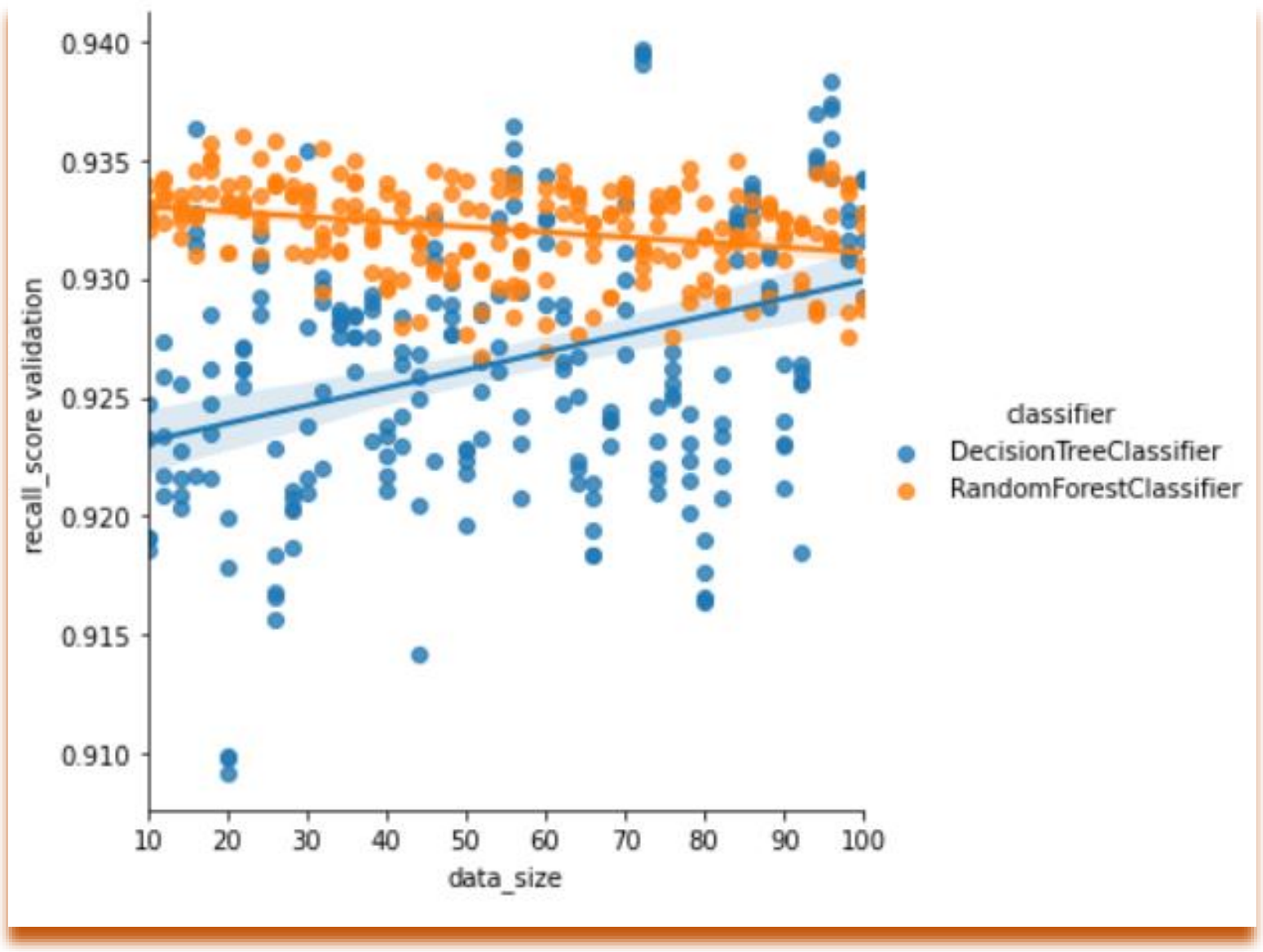

Figure 9. Recall score of the ML models of the experiment 1 (validation).

The precision scores (Fig. 12) of both models are in line with the results of the testing data (Fig. 7). Models have high accuracies even with as little as the ten percent of the total training data records. Both models' scores have improved when the number of training data has grown. The decision tree exhibits greater variance than the random forest model.

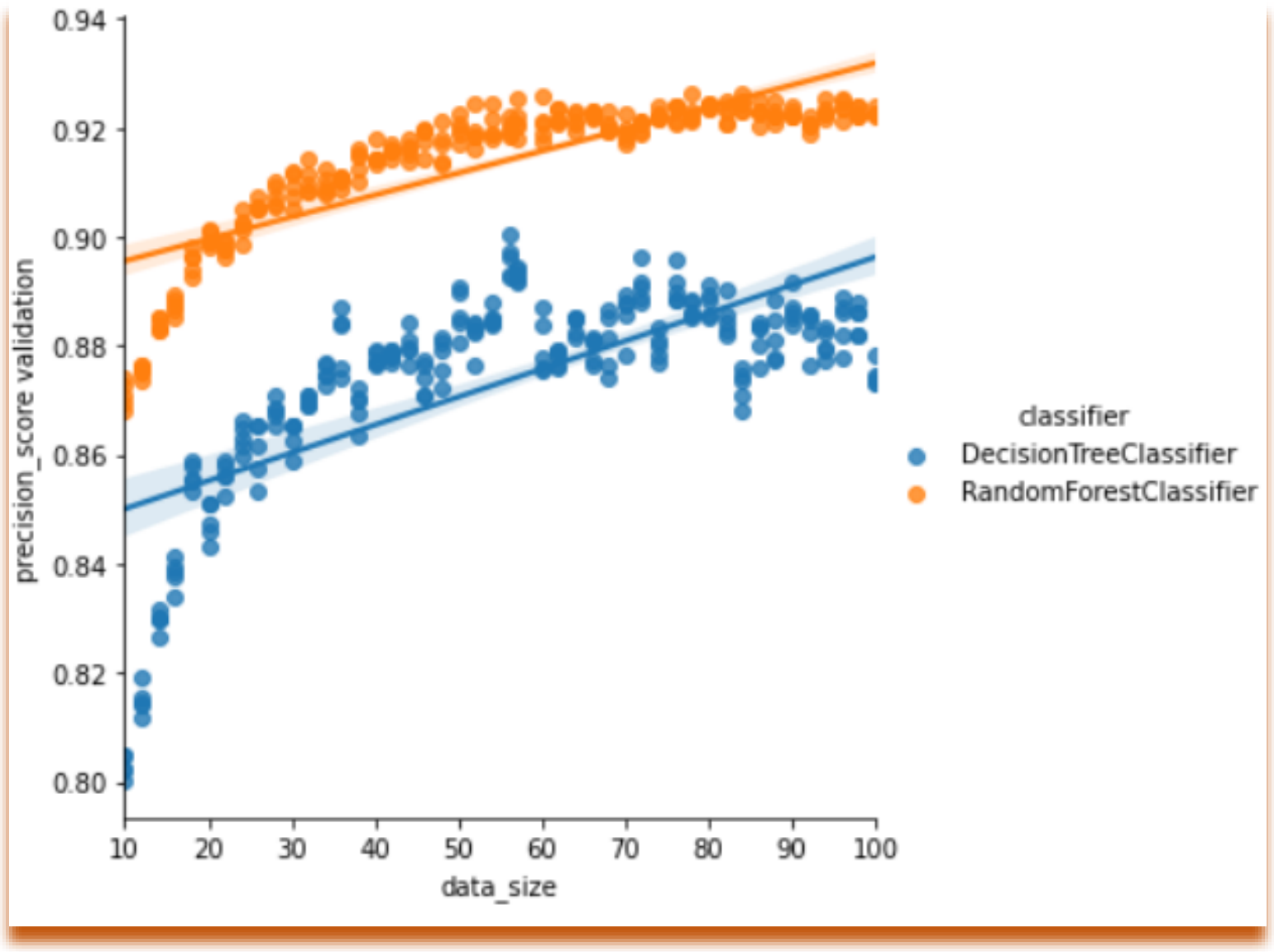

Figure 10. Precision score of the ML models of the experiment 1 (validation).

*B. Experiment 2 - addition of dimensions to data*

Visual inspection of the metrics used to measure the models' performances all the models gain better scores by these metrics when they have more features in their inputs. Utilizing just one feature leaves the scores low on every metric. However, the increase in the metrics when the number of input features grows can be very small between individual features. The ROC AUC score results of the experiment 2 are shown in the figures 13. The table 3 presents the mean, minimum, and maximum scores on all metrics for the decision tree classifiers. It clearly shows that adding more input features does not always significantly increase the metrics' scores. For example, the difference between means of the accuracies of 4th and 5th features (0.803166 and 0.803944, respectively) is 0.000778 which is less than the 2 percent limit generally thought as significant.

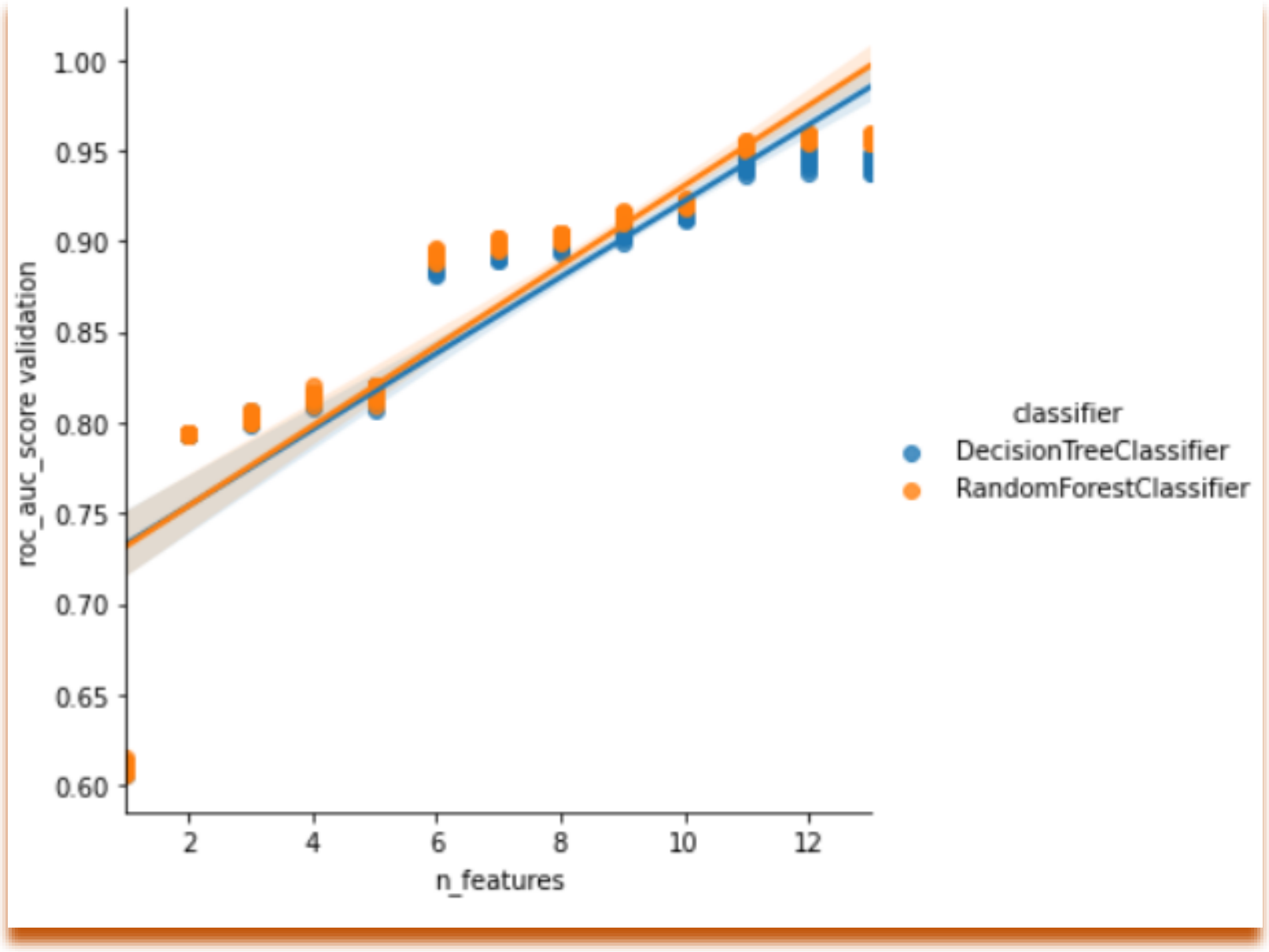

Figure 11. ROC AUC score of the ML models of the experiment 2 (validation).

TABLE III. METRIC SCORES OF THE DT MODELS OF EXPERIMENT 2

| DecisionTreeClassifier | Number of input features | 1 | 2 | 3 | 4 | 5 |
|---|---|---|---|---|---|---|
| **accuracy_score (validation)** | **mean** | **0.529855** | **0.766785** | **0.783992** | **0.803166** | **0.803944** |
| | min | 0.511772 | 0.765439 | 0.770395 | 0.785785 | 0.790293 |
| | max | 0.550265 | 0.767675 | 0.792293 | 0.808513 | 0.823500 |
| **roc_auc_score (validation)** | **mean** | **0.609651** | **0.793818** | **0.803154** | **0.813538** | **0.815594** |
| | min | 0.605324 | 0.792722 | 0.798690 | 0.807663 | 0.807152 |
| | max | 0.612589 | 0.794869 | 0.807182 | 0.815986 | 0.820431 |
| **jaccard_score (validation)** | **mean** | **0.239289** | **0.417431** | **0.435205** | **0.456875** | **0.459269** |
| | min | 0.236776 | 0.416257 | 0.423956 | 0.439774 | 0.449629 |
| | max | 0.241259 | 0.418744 | 0.443821 | 0.462040 | 0.472058 |
| **recall_score (validation)** | **mean** | **0.742302** | **0.838757** | **0.835008** | **0.830779** | **0.834960** |
| | min | 0.710654 | 0.836271 | 0.823619 | 0.823442 | 0.792131 |
| | max | 0.767426 | 0.841607 | 0.848213 | 0.845554 | 0.859912 |
| **precision_score (validation)** | **mean** | **0.261023** | **0.453851** | **0.476266** | **0.503828** | **0.505650** |
| | min | 0.256908 | 0.452238 | 0.458770 | 0.478674 | 0.485166 |
| | max | 0.265302 | 0.455015 | 0.487534 | 0.512086 | 0.538803 |

## V. CONCLUSION

In this paper we attempted to contemplate and answer the questions related to the quality over quantity in machine learning utilizing vocational school data in a classification setting. Two questions were proposed "what happens to the model's performance when more data is used in the learning phase of the model?" and "will increasing the number of input features improve the model's performance?". Two experiments were devised which tried to answer these questions, respectively. The authors would like to note that in this article testing data refers to the separate portion of records available in the training data while the validation data contains a selection of all the students not present in the training nor testing data sets.

In general, the results of the experiments were in line with the assumption that more data does improve the classification results as does the increase of input features. In addition, the choice of machine learning model does impact the results with the ensemble models giving better results in this case. In the experiment one, there were some peculiar results such as the poor recall score of the testing data and high recall result of the validation data indicating a problem in the experiment code or more precisely the correct use of the testing and validation data. With the experiment 2 it was shown that the increase of input features does improve the results if the input features themselves are in some ways significant for the data model. Thus, we can say the information gain via increasing the number of input features is case and feature specific. In addition, one should not use just one input feature, because adding a second input features immediately raised the scores significantly higher than just utilizing one input feature. However, one should note that the experiment 2 had used primary component analysis in the selection of the input features' order for the experiment thus the significant increase in the results will vary if the order of the input features for the experiment would be different. The other implementation choice would have been to permutate all the input features in decreasing order however that could have taken awhile. The authors agree that there are ways to enrich the single input feature to contain information from other features, such as utilizing the encoded output of an autoencoder, but the authors also argue that would make the interpretability of the models and the experiment set up more difficult. Testing enhanced single input features was out of the scope of the paper.

For the next research steps, the authors propose to implement a feature permutation method when selecting the input features, although it is of concern how long the implementation of the code would run with such choice. In addition, another design of the experiment one could be related to the number of unknown or missing values in the input features. Lastly, experimenting with utilizing just one input feature enhanced with autoencoders could be of interest